\documentclass{article}
\usepackage{spconf,graphicx}
\usepackage{amsmath, amssymb}
\usepackage{cite}

\usepackage{multirow}

\title{MVAFormer: RGB-based Multi-View Spatio-Temporal Action Recognition with Transformer}
\name{Taiga Yamane, Satoshi Suzuki, Ryo Masumura, Shotaro Tora}
\address{NTT Human Informatics Laboratories, NTT Corporation, Japan}
\begin{document}
%
\maketitle
\begin{abstract}

Multi-view action recognition aims to recognize human actions using multiple camera views and deals with occlusion caused by obstacles or crowds.
In this task, cooperation among views, which generates a joint representation by combining multiple views, is vital.
Previous studies have explored promising cooperation methods for improving performance.
However, since their methods focus only on the task setting of recognizing a single action from an entire video, they are not applicable to the recently popular spatio-temporal action recognition~(STAR) setting, in which each person's action is recognized sequentially.
To address this problem, this paper proposes a multi-view action recognition method for the STAR setting, called MVAFormer.
In MVAFormer, we introduce a novel transformer-based cooperation module among views.
In contrast to previous studies, which utilize embedding vectors with lost spatial information, our module utilizes the feature map for effective cooperation in the STAR setting, which preserves the spatial information.
Furthermore, in our module, we divide the self-attention for the same and different views to model the relationship between multiple views effectively.
The results of experiments using a newly collected dataset demonstrate that MVAFormer outperforms the comparison baselines by approximately $4.4$ points on the F-measure.
\end{abstract}

\begin{figure}[t]
    \centering
    \includegraphics[width=80mm]{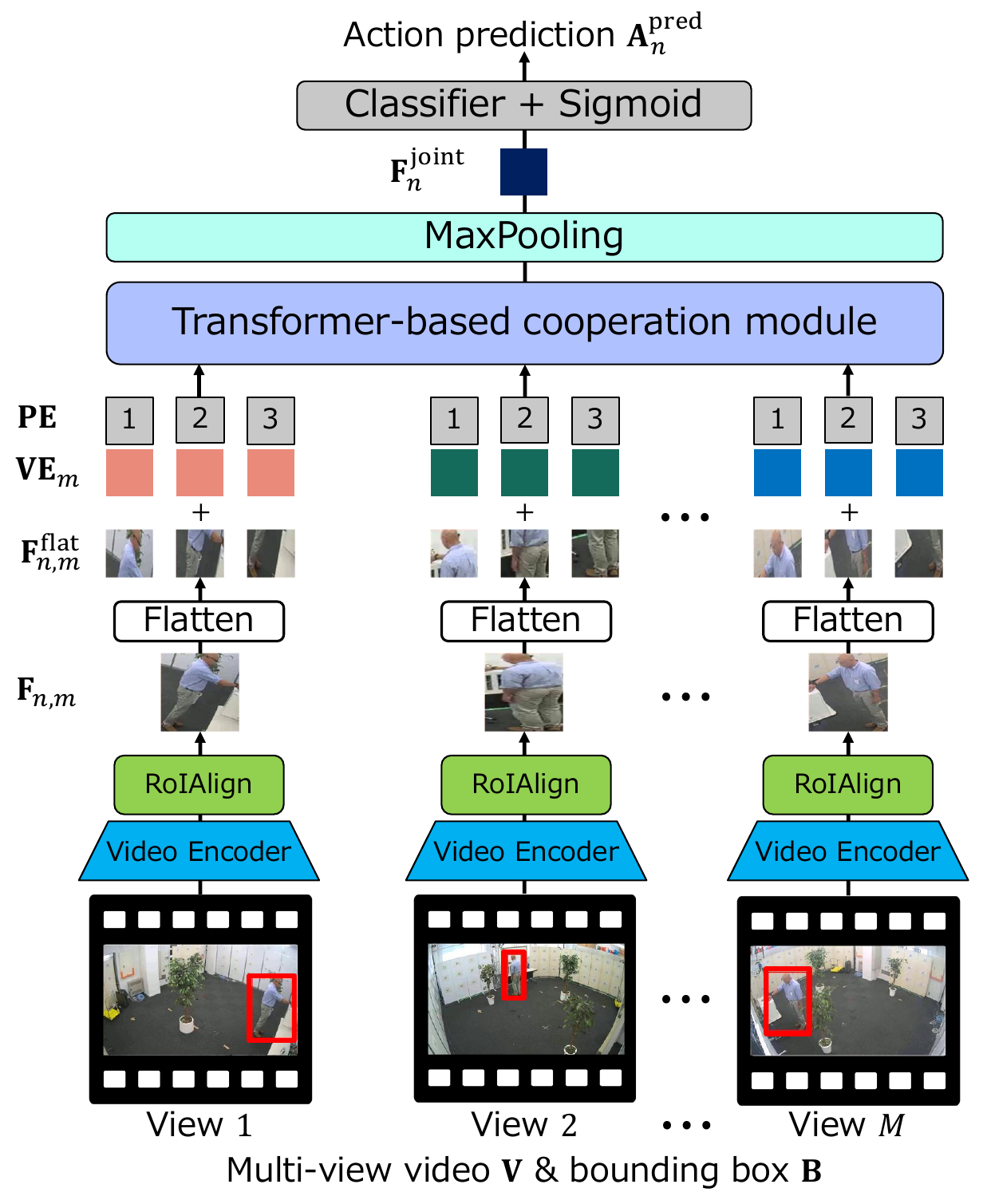}
    \vspace{-2mm}
    \caption{Overview of MVAFormer. For simplicity, we show an example where only one person is depicted in the multi-view video.}
    \vspace{-2mm}
    \label{fig:fig1}
\end{figure}

\section{Introduction}
\label{sec:intro}

Action recognition in videos~\cite{caba2015activitynet,carreira2017quo} is one of the most studied fields within the computer vision community, and many researchers are continually exploring increasingly practical task settings.
For example, in early studies~\cite{carreira2017quo,goyal2017something}, the primary goal was to recognize a single action in an entire video, assuming that only one action exists in a video.
This setting is not applicable in scenarios involving multiple persons taking successive different actions, e.g., in surveillance systems and healthcare services where the recognition of each action is required.
Therefore, recent studies have shifted their focus to the spatio-temporal action recognition~(STAR) setting~\cite{gu2018ava,li2020ava}.
In this setting, the goal is to recognize the actions of each person within a video at specific time intervals.
The STAR setting is more challenging than the earlier setting because it needs to cover predictions that are spatially and temporally more dense in both directions.

In the early and STAR setting, deep learning-based methods have recently emerged as the dominant choice and have showcased outstanding performance~\cite{carreira2017quo,feichtenhofer2019slowfast,dosovitskiy2021image,tong2022videomae}.
However, these methods are known to be vulnerable to occlusion caused by obstacles or crowds~\cite{kong2019mmact}, leading to a decline in performance in such scenarios, as they rely solely on a single RGB view.
To overcome this problem, multi-view action recognition, which is a task to recognize actions using multiple views from different angles, has been attracting interest~\cite{shahroudy2016ntu,kong2019mmact}.
Utilizing views from multiple angles can significantly reduce occluded areas and may achieve satisfactory performance in scenarios involving obstacles or crowds. 
The key to multi-view action recognition lies in cooperation among views, which generates a good joint representation from multiple views by finding and integrating dispersed cues in each view. 
Therefore, many studies have focused on the development of cooperation methods~\cite{chen2021channel,friji2021geometric,wang2018dividing,yasuda2022multi}.

In multi-view action recognition, one of the most widely studied cooperation approaches is skeleton-based cooperation~\cite{chen2021channel,friji2021geometric}, which leverages the geometric relationships between joints of the skeleton.
As these approaches cannot use information outside the skeleton, they are limited in the actions they can recognize, including actions involving interactions with objects or other persons.
Another approach is RGB-based cooperation~\cite{wang2018dividing,yasuda2022multi}, which does not require 3D skeletons and only utilizes RGB views.
This approach is expected to recognize actions accurately, even in occlusion scenes, while maintaining high feasibility.
However, the previous works on RGB-based cooperation focus only on the early task setting of recognizing a single action from an entire video.
Therefore, their effectiveness for the STAR setting is limited because, unlike the early task setting, this setting needs to recognize many individual actions of multiple people, and the required cooperation method will thus be significantly different.

In this paper, we propose a novel RGB-based method for multi-view action recognition in the \textit{STAR setting}, called \textbf{MVAFormer}.
Fig.~\ref{fig:fig1} shows an overview of MVAFormer.
Its two main components are RoIAlign and the transformer-based cooperation module.
RoIAlign aims to extract each person's representation from each view.
We utilize it to adapt MVAFormer to the STAR setting, where each person's recognition is required.
The transformer-based cooperation module generates a good joint representation from multiple views using the representation extracted by RoIAlign as input.
In MVAFormer, we design a novel cooperation module for the STAR setting, as follows.
(i) Our cooperation module generates a joint representation using each person's feature map while preserving spatial information.
This is because utilizing spatial information is essential for effective cooperation among views in spatially dense recognition tasks~\cite{hou2020multiview,wang2022detr3d}.
In contrast, the previous studies~\cite{wang2018dividing,yasuda2022multi} use the embedding vector, which has no spatial information obtained from the video encoder, since they only focus on the early task setting.
(ii) In the transformer, we divide the self-attentions into the same view attention and different view attention~(Fig.~\ref{fig:fig2}), unlike the vanilla transformer~\cite{vaswani2017attention,yasuda2022multi}, which uses only the self-attention.
This enables the transformer to effectively model the relationship between multiple views by forcing the attention to different views.

To the best of our knowledge, there is no multi-view dataset for the STAR setting.
Therefore, we constructed a new dataset based on the MMAct~\cite{kong2019mmact} and AVA~\cite{gu2018ava} datasets to verify the effectiveness of MVAFormer in the STAR setting.
Our dataset contains videos of occlusion scenes captured from four synchronized RGB cameras and annotations consisting of action labels and bounding boxes for the STAR setting.
The results of experiments conducted using our dataset demonstrate that the proposed method outperforms the comparison methods by $5.5$ points on precision, $3.2$ points on recall, and $4.4$ points on F-measure.

\section{Related Work}
\label{sec:related}

\subsection{Action Recognition}
\label{subsec:singleview}

There are many task settings for action recognition~\cite{carreira2017quo,gu2018ava}.
The most studied task setting is recognizing a single action throughout a video~\cite{carreira2017quo,goyal2017something}.
In this task setting, the main research focus has been effective feature extraction from the video to encode appearance and motion information~\cite{zhu2020comprehensive}.
Traditional methods have relied on hand-crafted features~\cite{turaga2008machine}, but following the success of AlexNet~\cite{krizhevsky2012imagenet} in image classification, deep learning-based video encoders have achieved state-of-the-art performance in action recognition~\cite{carreira2017quo,feichtenhofer2019slowfast,dosovitskiy2021image,tong2022videomae}.
\cite{carreira2017quo,feichtenhofer2019slowfast} effectively extract spatial and temporal information by utilizing 3D CNN, an extension of 2D CNN to the time axis utilized in image recognition.
\cite{dosovitskiy2021image,tong2022videomae} modified a transformer, which had been successful in natural language processing~\cite{vaswani2017attention}, for action recognition and achieved better performance than the 3D CNN methods.
However, all methods in the above setting have limited application because they cannot deal with multiple people and sequential actions.

In recent years, the spatio-temporal action recognition~(STAR) task setting has received significant attention due to its practicality~\cite{gu2018ava,li2020ava}.
STAR aims to recognize each person's action sequentially in scenarios involving multiple people taking successive actions from the video and bounding boxes that represent people's locations.
The most famous dataset for the STAR setting is AVA~\cite{vaswani2017attention}, which contains actions and bounding boxes annotated on keyframes at one-second intervals, and this dataset has greatly advanced the research on the STAR setting.
Several studies have proposed methods specifically designed for the STAR setting~\cite{pan2021actor,feng2021relation}, but a simple method that adds RoIAlign~\cite{ren2015faster} after the video encoder has achieved a state-of-the-art result~\cite{feichtenhofer2019slowfast,tong2022videomae}.
This simple method extracts a representation of each person from a video using RoIAlign with pre-computed or ground truth bounding boxes for spatially and temporally dense action recognition.
While these methods have achieved excellent performance on the benchmark datasets, they often suffer a drop in performance in scenes with occlusion, as they utilize only a single RGB view~\cite{kong2019mmact}.

\subsection{Multi-View Action Recognition}
\label{subsec:multiview}

Multi-view action recognition has recently attracted interest as a way of mitigating the effect of occlusion, and various datasets and methods have been proposed~\cite{shahroudy2016ntu,kong2019mmact}.
NTU-RGB+D~\cite{shahroudy2016ntu} is the most used dataset in multi-view action recognition, consisting of not only multi-view RGB videos but also depth and 3D skeleton data, but the videos in the NTU-RGB+D are only captured in scenes with little occlusion.
MMAct~\cite{kong2019mmact} is a dataset aimed at accurately recognizing actions in scenes with high occlusion.
It consists of multi-view RGB videos captured by four cameras installed in the corners of a room and other modality data such as acceleration and pressure signals.
Since only one action label is annotated to one multi-view video in these datasets, they cannot be applied directly to the STAR setting.
To the best of our knowledge, there is no multi-view dataset for the STAR setting.
Therefore, we construct a new dataset based on MMAct.

For effective multi-view action recognition, the key to success lies in cooperation among views, which generates a good joint representation from multiple views.
This is because the cues for actions are dispersed across the views, and the appearance of the person is different in each view.
\cite{wang2018dividing} introduced a conditional random field~(CRF) to exchange information among views from embedding vectors obtained from each view by the video encoder.
In \cite{yasuda2022multi}, embedding vectors representing each view are utilized by a transformer to effectively cooperate among views.
However, these methods are unable to recognize the spatially and temporally dense actions required in the STAR setting, since they extract an embedding vector that represents the entire video of a single view and cooperate among views based on this embedding vector.
Furthermore, because these methods use vectors with lost spatial resolution for cooperation, they do not fully exploit the spatial information in each view that is important for spatially dense multi-view recognition~\cite{hou2020multiview,wang2022detr3d}.

\section{Proposed Method}
\label{sec:method}

\subsection{Problem Setting}
\label{subsec:problem}

In this paper, we consider the case of recognizing the actions of $N$ people at one keyframe in a multi-view video captured by synchronized $M$ cameras.
Let us denote the multi-view video as $\mathbf{V} = \{ \mathbf{V}_m \}^{M}_{m=1}$, where $\mathbf{V}_m \in \mathbb{R}^{T \times H \times W \times 3}$ is the video from the $m$-th view.
Here, $T$, $H$, and $W$ are the temporal, height, and width dimensions of the video of each view, respectively.
The bounding box $\mathbf{B} = \{ ( \mathbf{b}_{1, m}, \cdots, \mathbf{b}_{n, m}, \cdots,$ $\mathbf{b}_{N, m} ) |^{M}_{m=1} \}$ is annotated to the middle frame (i.e., keyframe) of $\mathbf{V}$, where $\mathbf{b}_{n, m} \in \mathbb{R}^{4}$ represents the $n$-th person's box coordinates in the $m$-th view.
$n$ is the person index, and the same $n$ indicates the same person.
The $\mathbf{B}$ is obtained by human annotation or preprocessing such as person detector ~\cite{ren2015faster,carion2020end}.
When the $n$-th person is not in the $m$-th view due to occlusion or other reasons, $\mathbf{b}_{n, m}$ becomes an empty set $\varnothing$.
In this task, we need to predict the presence or absence of each action $\mathbf{A} = \{ \mathbf{A}_n \}^{N}_{n=1}$ using $\mathbf{V}$ and $\mathbf{B}$, where $\mathbf{A}_n \in \{0, 1\}^\mathrm{cls}$ represents $n$-th person's action, and $\mathrm{cls}$ is the number of predefined actions.
This prediction process is performed for all keyframes in a multi-view video.

\subsection{MVAFormer}
\label{subsec:mvaformer}

Fig.~\ref{fig:fig1} shows an overview of our proposed MVAFormer, which mainly consists of the shared video encoder, RoIAlign, and transformer-based cooperation module.
The video encoder extracts the feature from each view video $\mathbf{V}_m$, and the RoIAlign then extracts each person's representation from each view by cutting out the region defined by the bounding box $\mathbf{B}$ at a fixed size from that feature.
We set the spatial resolution of output from RoIAlign to $7 \times 7$.
To cooperate with the outputs of RoIAlign among views, we adopt the transformer~\cite{vaswani2017attention}, since it has been reported to be effective for cooperation among multiple sensors~\cite{sun2019videobert,yasuda2022multi}.

The formulations of MVAFormer are as follows.
First, the video encoder and RoIAlign extract each person's representation $\{ ( \mathbf{F}_{1, m}, \cdots, \mathbf{F}_{n, m}, \cdots, \mathbf{F}_{N, m} ) |^{M}_{m=1} \}$ from input videos $\mathbf{V}$ and bounding box $\mathbf{B}$, as
\begin{align}
    \centering
    \mathbf{F}_{n, m} = \mathrm{RoIAlign}(\mathrm{VideoEncoder}(\mathbf{V}_m), \mathbf{b}_{n, m}),
\label{formula:1}
\end{align}
where $\mathbf{F}_{n, m} \in \mathbb{R}^{7 \times 7 \times c}$ is the $n$-th person's representation in the $m$-th view and $c$ is the channel dimension.
When the $n$-th person is not in the $m$-th view~(i.e., $\mathbf{b}_{n, m} = \varnothing$), $\mathbf{F}_{n, m}$ is padded by $\mathbf{0}$.
The video encoder is shared by all views and can be applied to any models for single-view action recognition, as described in Sec.~\ref{subsec:singleview}.

Then, the transformer-based cooperation module generates a joint representation for each person.
Our MVAFormer cooperates among views by directly utilizing the feature map $\mathbf{F}_{n, m}$ that preserves spatial resolution.
Conventional cooperation methods~\cite{wang2018dividing,yasuda2022multi} first perform the global pooling on each person's feature map $\mathbf{F}_{n, m}$ to obtain the embedding vector, and then generate a joint representation based on that vector.
However, since this embedding vector loses spatial resolution, these methods do not fully utilize the spatial information that is important for effective cooperation in spatially dense recognition tasks~\cite{hou2020multiview,wang2022detr3d}.

To enable the transformer to handle a 2D feature map, we flatten $\mathbf{F}_{n, m} \in \mathbb{R}^{7 \times 7 \times c}$ into $\mathbf{F}_{n, m}^{\mathrm{flat}} \in \mathbb{R}^{(7 \cdot 7) \times c}$.
We need to provide the position information and view index information to $\mathbf{F}_{n, m}^{\mathrm{flat}}$ before it is input to the transformer because the transformer does not distinguish the order of the input sequences.
To provide the position information to $\mathbf{F}_{n, m}^{\mathrm{flat}}$, we adopt the fixed sinusoidal position embeddings $\mathbf{PE} \in \mathbb{R}^{(7 \cdot 7) \times c}$, the same as \cite{carion2020end}. 
$\mathbf{PE}$ is shared by all $\mathbf{F}_{n, m}^{\mathrm{flat}}$.
Unlike each pixel in $\mathbf{F}_{n, m}$, the feature maps from different view (i.e., $( \mathbf{F}_{n, 1}^{\mathrm{flat}}, \cdots, \mathbf{F}_{n, M}^{\mathrm{flat}} )$) do not have a continuous relationship.
Therefore, we introduce learnable view embeddings $ \{ \mathbf{VE}_m \}^{M}_{m=1}$ to provide view index information, where $\mathbf{VE}_m \in \mathbb{R}^{c}$ is the view embedding of the $m$-th view.
After adding position embeddings and view embeddings to $\mathbf{F}_{n, m}^{\mathrm{flat}}$, we generate the joint representation $\{ \mathbf{F}_{n}^{\mathrm{joint}} \}^{N}_{n=1}$ utilizing the transformer as follows,
\begin{align}
    \centering
    \vspace{-10mm}
    \mathbf{\tilde{F}}_{n, m}^{\mathrm{flat}} &= \mathbf{F}_{n, m}^{\mathrm{flat}} + \mathbf{PE} + \mathbf{VE}_{m}, \\
    \mathbf{F}^{\mathrm{joint}}_n &= \mathrm{MaxPooling}(\mathrm{Transformer}(\{ \mathbf{\tilde{F}}_{n, m}^{\mathrm{flat}} \}^{M}_{m=1})),
\label{formula:2}
\end{align}
where $\mathbf{F}^{\mathrm{joint}}_n \in \mathbb{R}^{c}$ is the $n$-th person's joint representation.
Finally, a linear classifier and sigmoid function are used to output  $\mathbf{A^{\mathrm{pred}}} = \{ \mathbf{A}_n^{\mathrm{pred}} \}^{N}_{n=1}$, where $\mathbf{A}_n^{\mathrm{pred}}$ is the action prediction for the $n$-th person.

\begin{figure}[t]
    \centering
    \includegraphics[width=75mm]{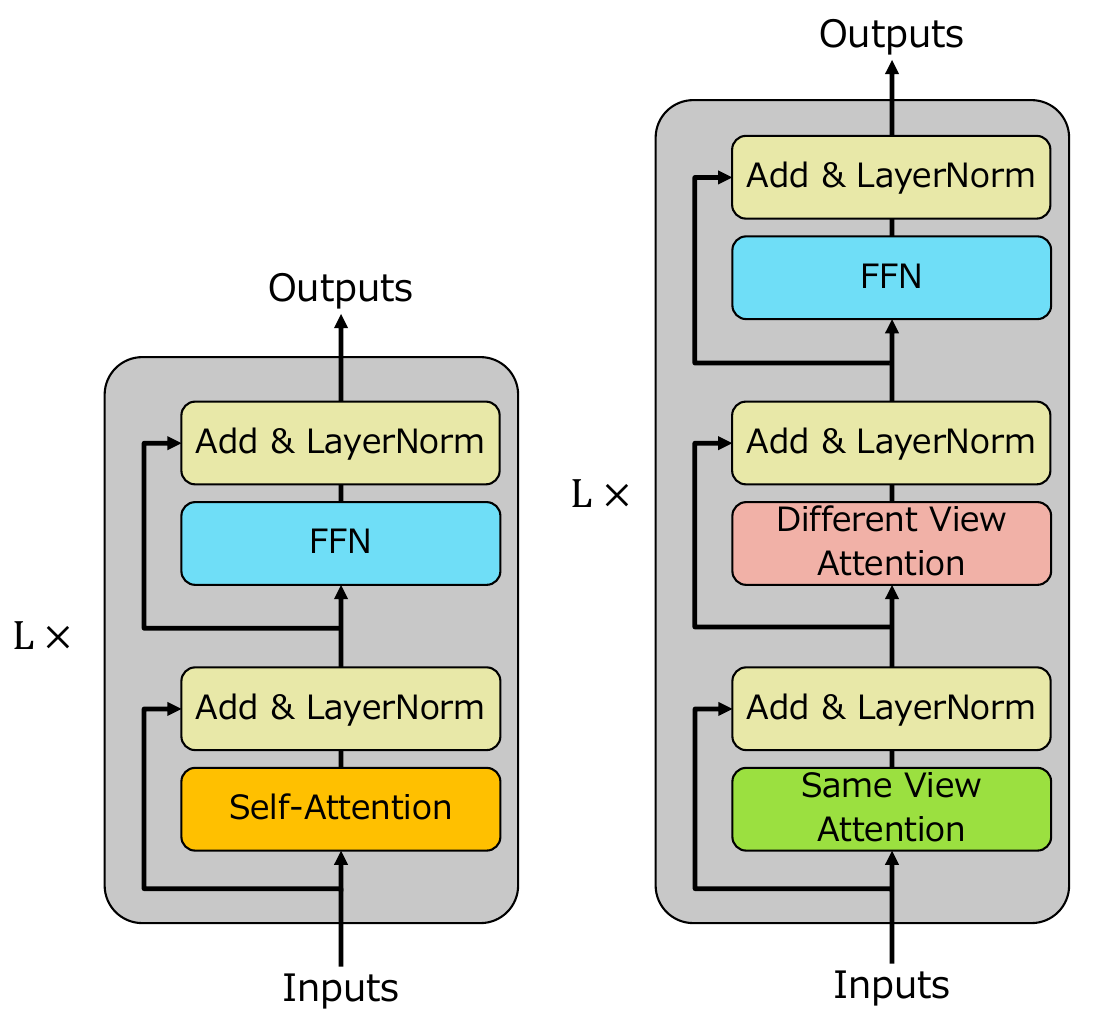}
    \vspace{-2mm}
    \caption{(\textbf{Left}) Vanilla transformer, which consists of Self-Attention, FFN, and LayerNorm. (\textbf{Right}) Our transformer, which consists of the Same View Attention~(SVA), Different View Attention~(DVA), FFN, and LayerNorm.}
    \vspace{-2mm}
    \label{fig:fig2}
\end{figure}

\subsection{Dividing Self-Attention}
\label{subsec:dividing}

\begin{figure}[t]
    \centering
    \includegraphics[width=80mm]{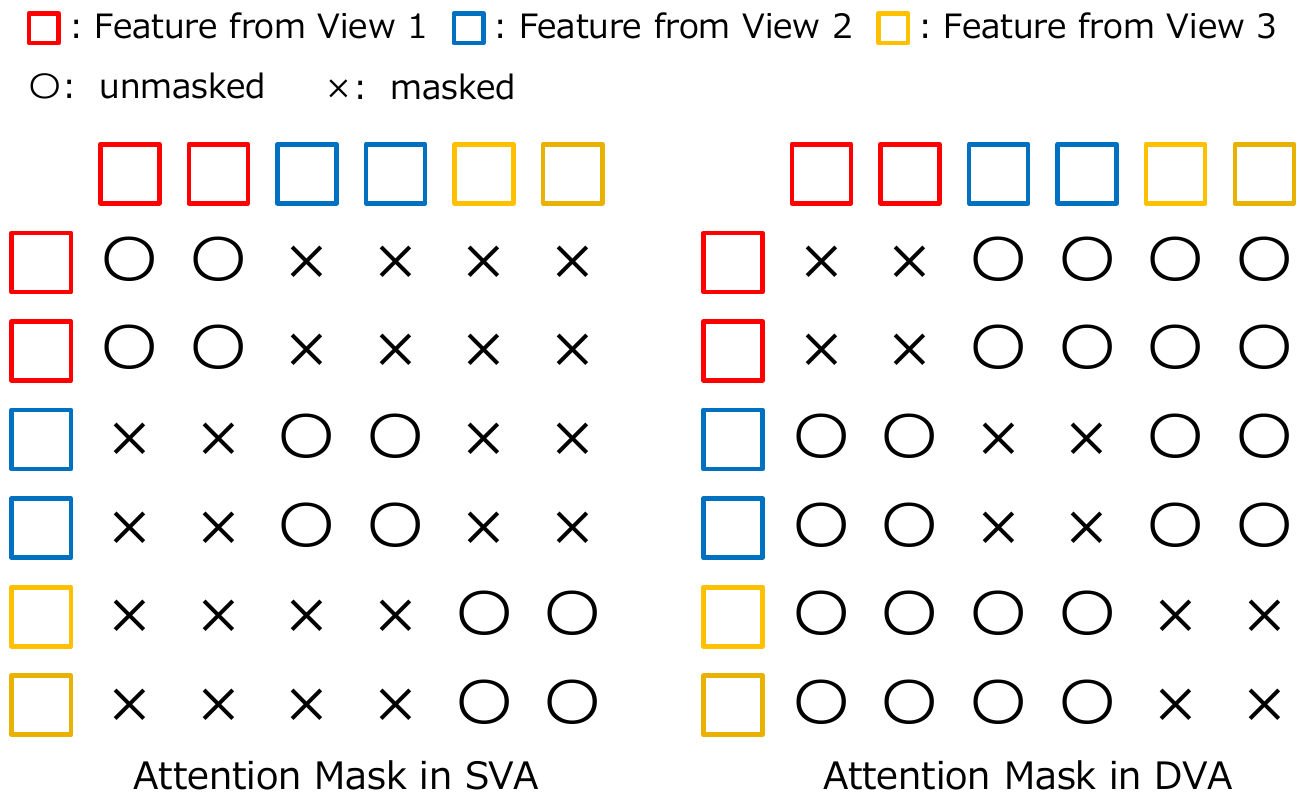}
    \vspace{-2mm}
    \caption{Attention mask in (\textbf{Left}) SVA and (\textbf{Right}) DVA. Each row indicates the query in the attention. Each column indicates the key and value in the attention.}
    \vspace{-2mm}
    \label{fig:fig3}
\end{figure}

Fig.~\ref{fig:fig2} shows the vanilla transformer~\cite{vaswani2017attention} and our transformer.
As shown on the left, the vanilla transformer has $L$ layers, each of which consists of self-attention, feed-forward network~(FFN), and layer normalization~(LayerNorm).
This self-attention extracts the relationship among the input sequences based on the pairwise similarity between two elements of the sequences.
As shown on the right, we divide the self-attention into same view attention~(SVA) and different view attention~(DVA) for more effective relation extraction among views than the self-attention.
The self-attention in computer vision has been reported that attention is concentrated on features with high similarity of the same object~\cite{carion2020end}.
This phenomenon causes relationship extraction by the self-attention to concentrate excessively on features in the same view with high similarity, resulting in inadequate relationship extraction between different views.
To avoid this concentration, the SVA extracts the relationship only between features of the same view and the DVA extracts the relationship only between different views, as
\begin{align}
    \centering
    &\mathbf{F}_{n, m}^{\mathrm{SVA}} = \mathrm{Attention}(\mathbf{\tilde{F}}_{n, m}^{\mathrm{flat}}, \mathbf{\tilde{F}}_{n, m}^{\mathrm{flat}}, \mathbf{\tilde{F}}_{n, m}^{\mathrm{flat}}), \\
    &\mathbf{F}_{n, m}^{\mathrm{DVA}} = \mathrm{Attention}(\mathbf{\tilde{F}}_{n, m}^{\mathrm{flat}}, \{\mathbf{\tilde{F}}_{n, l}^{\mathrm{flat}}\}_{l \ne m}, \{\mathbf{\tilde{F}}_{n, l}^{\mathrm{flat}}\}_{l \ne m}),
\label{formula:3}
\end{align}
where $\mathbf{F}_{n, m}^{\mathrm{SVA}}$ and $\mathbf{F}_{n, m}^{\mathrm{DVA}}$ are the $n$-th person's $m$-th view outputs from SVA and DVA, respectively.
Here, the first, second, and third arguments of the above $\mathrm{Attention}$ function are query, value, and key in the attention mechanism, respectively.
Specifically, the SVA and DVA are implemented by combining the self-attention and attention masks~\cite{vaswani2017attention}.
The strategy of attention masks in the SVA and DVA are shown in Fig.~\ref{fig:fig3} left and right, respectively, where $\circ$ and $\times$ respectively indicate unmasked and masked.
On the left, because $\circ$ is displayed only between features in the same view~(same color), the SVA is forced to attend only to the feature from the same view.
On the right, because $\circ$ is displayed only between features in different views~(different colors), the DVA is forced to attend only to the feature from different views.

\begin{figure*}[tb]
    \centering
    \includegraphics[width=150mm]{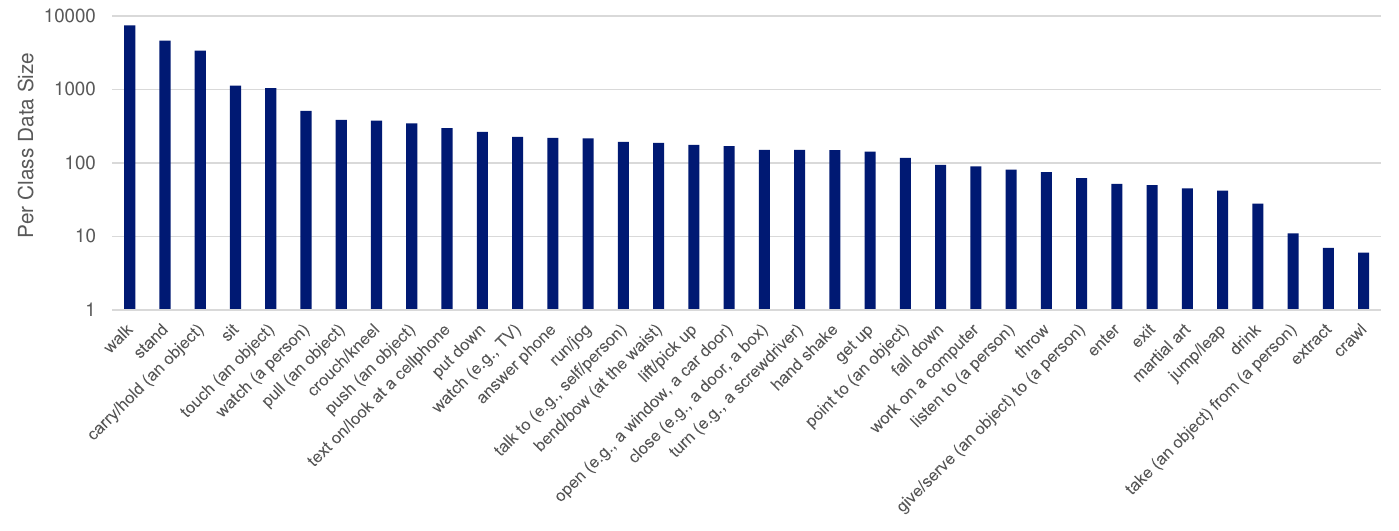}
    \vspace{-5mm}
    \caption{Size of each action class in our dataset in descending order.}
    \vspace{-2mm}
    \label{fig:fig4}
\end{figure*}

\begin{table*}[tb]
\centering
\vspace{-3.0mm}
\caption{Results of comparison with other methods. Bold type indicates the highest performance.
}
\begin{tabular}{c|l|ccc|c} \hline
Input & Method & Precision & Recall & F-measure & FLOPs \\ \hline \hline
\multirow{2}{*}{Single-view} & Predicting on only one view     & 0.667   & 0.560  & 0.588 & 180G\\ 
                              & Ensemble of each view          & 0.753   & 0.537  & 0.597 & 720G \\ \hline
\multirow{4}{*}{Multi-view}  & CRF~\cite{wang2018dividing}     & 0.750   & 0.570  & 0.618 & 721G \\
                              & MultiTrans~\cite{yasuda2022multi} & 0.713   & 0.594  & 0.621 & 722G\\
                              & MVAFormer w/o SVA \& DVA       & 0.761    & 0.595 & 0.638 & 735G\\ 
                              & MVAFormer                      & \textbf{0.768} & \textbf{0.626} & \textbf{0.665} & 739G\\ \hline
\end{tabular}
\vspace{-2.0mm}
\label{table:result}
\end{table*}

\section{Experiments}
\label{sec:experiment}

\subsection{Dataset}
\label{subsec:dataset}

To the best of our knowledge, there is no multi-view action recognition dataset that fits the problem setting described in Sec~\ref{subsec:problem}.
Therefore, to verify the effectiveness of MVAFormer, we construct a new dataset, in which the multi-view videos of the MMAct~\cite{kong2019mmact} are annotated with human actions and bounding boxes, the same as the AVA~\cite{gu2018ava}.

The MMAct provides multi-view videos captured by four cameras in the corners of a room for four different scenes, which have various types of obstacles.
Therefore, the MMAct fits our problem setting in terms of multi-view video data.
However, the MMAct is not annotated for the STAR setting.
For the single-view STAR setting, the AVA is the most famous dataset.
In the AVA, the bounding box and action labels are annotated to keyframes at one-second intervals for videos captured by one camera.

From the above two datasets, we construct a new dataset by annotating the bounding box and action labels as in the AVA to the multi-view videos of the MMAct.
Following the same annotation method as AVA, we collected $14$k annotations, which included $36$ of AVA's $80$ action classes.
Fig.~\ref{fig:fig4} shows the distribution of action annotations in our dataset.
In the experiments, we split the dataset into $11$K for training and $3$K for evaluation with the same class distribution and no person overlapping.
Since the distribution of our dataset was long-tailed, the bottom three action classes, which are extremely small in number, were excluded in the evaluation.

\subsection{Implementation Details}
\label{subsec:imple}

Videos in the MMAct dataset are captured by four cameras, so we set $M=4$.
The video input was downsampled from $30$ to $7.5$ fps and the resolution was compressed to $224 \times 224$.
Unless otherwise mentioned, for all experiments, we used 16-frame ViT-Base~\cite{dosovitskiy2021image} pretrained with VideoMAE~\cite{tong2022videomae} as the video encoder.
We set the number of transformer layers to $L=4$ and the number of attention heads to $4$.
To investigate the upper bound of the performance of MVAFormer, we used ground truth bounding boxes for all training and evaluation.

For all training, we set the batch size to 128 and the epoch to 25.
We used BCE loss and the AdamW optimizer~\cite{loshchilov2018decoupled}, set the initial learning rate to $1.0 \times 10^{-4}$ and decayed to $1.0 \times 10^{-6}$ following a cosine schedule.

To evaluate the effectiveness of our MVAFormer, we compare it with existing single-view and multi-view methods.
As single-view methods, we implemented two methods as baselines.
The first is to predict for only one of the four views in the same way as for the single-view STAR settings~\cite{tong2022videomae}.
The second is the mean ensemble from the predictions of each view.
Since these single-view methods do not utilize relationships among views, we verify the necessity of cooperation among views in the STAR setting by comparing the MVAFormaer method with these single-view methods.
As multi-view methods, we adopt CRF~\cite{wang2018dividing} and MultiTrans~\cite{yasuda2022multi}, which cooperate among views using embedding vectors with lost spatial information.
These multi-view methods recognize only one action for one video, as discussed in Sec.~\ref{subsec:multiview}, so we modified them by adding RoIAlign and MaxPooling after the video encoder to enable the recognition of spatially and temporally dense actions.
By comparing our MVAFormer with these multi-view methods, we verify the effectiveness of our proposed cooperation methods.

As evaluation metrics, we adopt precision, recall, and F-measure, which are widely adopted in action recognition.
After calculating these metrics for each class, the score averaged over all classes is used as the overall score.

\subsection{Results}
\label{subsec:results}

\begin{table}[tb]
\centering
\vspace{-2.0mm}
\caption{Effect of dividing self-attention into SVA and DVA. ''Param'' indicates the number of parameters of the transformer. Bold type indicates the highest performance.
}
\scalebox{0.93}{
\begin{tabular}{c|cc|ccc} \hline
Attention type                  &  $L$  & Param  & Precision      & Recall         & F-measure\\ \hline\hline
\multirow{3}{*}{Self-Attention} &  4  &  37M   & 0.761          & 0.595          & 0.638\\ 
                                &  5  &  47M   & 0.763          & 0.588          & 0.636\\  
                                &  6  &  56M   & \textbf{0.775} & 0.585          & 0.638\\ \hline
SVA and DVA        &  4  &  47M   & 0.768          & \textbf{0.626} & \textbf{0.665}\\ \hline
\end{tabular}
}
\vspace{-2.0mm}
\label{table:attention}
\end{table}

\begin{figure}[t]
    \centering
    \includegraphics[width=75mm]{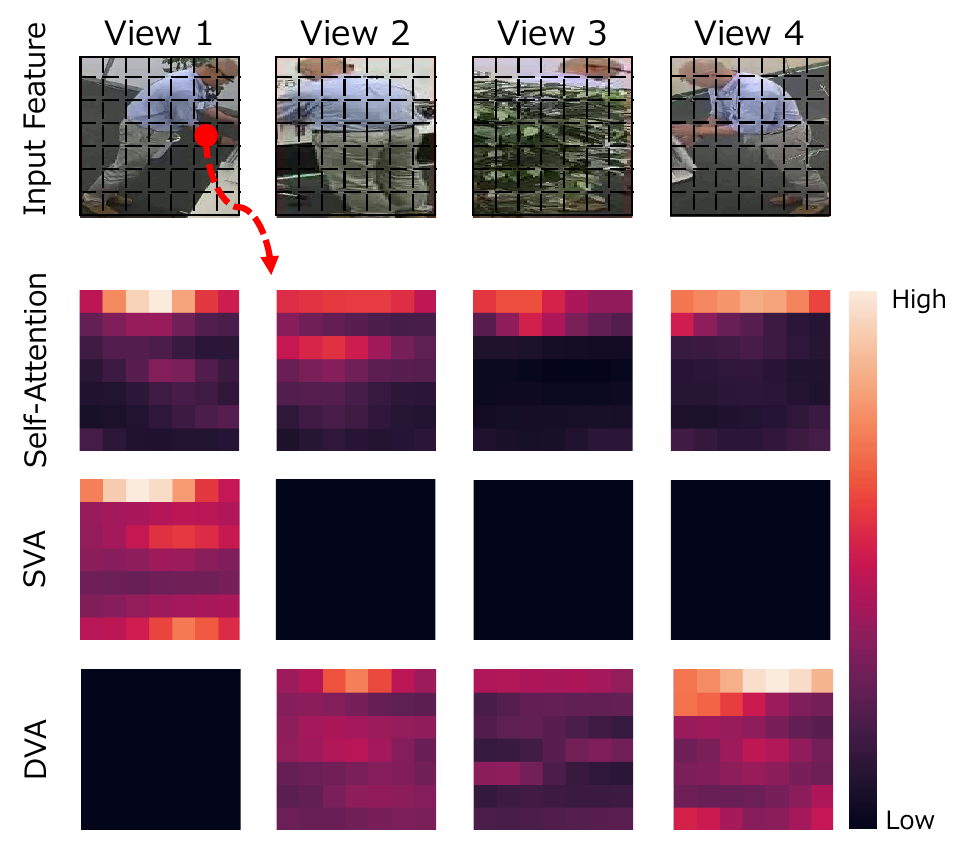}
    \vspace{-2mm}
    \caption{Visualization of attention weights for the red-filled circle feature from view $1$.}
    \vspace{-2mm}
    \label{fig:fig5}
\end{figure}

Table~\ref{table:result} shows the comparison results.
Our MVAFormer achieved higher performance than the single-view methods on all evaluation metrics.
Moreover, the other multi-view methods also achieved a higher performance than the single-view methods on almost all evaluation metrics.
This shows the necessity of cooperation among views in the STAR setting.
Among the multi-view methods, our MVAFormer without SVA and DVA outperformed CRF~\cite{wang2018dividing} and MultiTrans~\cite{yasuda2022multi} on all metrics, which cooperate among views utilizing embedding vectors with lost spatial information.
This indicates that spatial information is important for cooperation among views and that utilizing feature maps with spatial resolution leads to better cooperation.
In addition, by using SVA and DVA, MVAFormer achieved the highest performance of all, and the improvement from MultiTrans was $5.5$ points on precision, $3.2$ points on recall, and $4.4$ points on F-measure.

\subsection{Analysis}
\label{subsec:analysis}

The number of transformer parameters increases by dividing the self-attention into SVA and DVA.
To show that the performance improvement is not simply due to this parameter increase, we experimented with different numbers of transformer layers $L$ using self-attention.
Table~\ref{table:attention} shows the comparison results of $L=4, 5, 6$.
Increasing the number of the transformer parameters by larger $L$ improved precision but dropped recall.
In contrast, SVA and DVA improved all evaluation scores, with a particularly large increase in recall.

Next, to confirm the effectiveness of dividing the self-attention, we visualized the attention weights in the transformer.
Fig.~\ref{fig:fig5} provides an example of a visualization of the scene in which a person opens a locker, and shows the attention weights for the red-filled circle.
From the left to right represents the person's feature map from views $1$, $2$, $3$, and $4$.
From the top to bottom represents the input feature map to the transformer, attention weights in self-attention, SVA, and DVA.
In the self-attention, we can see that the occlusion area in the third view has lower attention weights than the area without occlusion.
This indicates that the transformer cooperates among views by adaptively selecting information that is useful for recognizing actions.
However, the attention weights in self-attention are most concentrated on the feature from the same view, as described in Sec.~\ref{subsec:dividing}.
By dividing self-attention, the attention weights in DVA are higher than self-attention in different views, while maintaining the ability of adaptive selection, as shown in the bottom right.
This indicates that dividing self-attention satisfies our objective of avoiding excessive concentration of relationship extraction within the same view, resulting in better performance.

To investigate the effect of the video encoder size, we experimented with four different sizes of ViT as the video encoder: Small~($22$M parameters), Base~($87$M), Large~($305$M), and Huge~($633$M).
Table~\ref{table:enc} shows the comparison results.
Increasing the size of the video encoder improves all evaluation scores.
By using the largest ViT-Huge, we achieved $0.824$ on precision, $0.646$ on recall, and $0.701$ on F-measure.

\begin{table}[tb]
\centering
\vspace{-2.0mm}
\caption{Effect of video encoder size. ''Param'' indicates the number of parameters of the video encoder. Bold type indicates the highest performance.
}
\scalebox{0.93}{
\begin{tabular}{c|c|ccc} \hline
Video encoder & Param  & Precision  & Recall & F-meaure\\ \hline\hline
ViT-Small     & 22M  & 0.751 & 0.508 & 0.571\\ 
ViT-Base      & 87M  & 0.768 & 0.626 & 0.665\\ 
ViT-Large     & 305M & 0.790 & 0.630 & 0.676\\ 
ViT-Huge      & 633M & \textbf{0.824} & \textbf{0.646} & \textbf{0.701}\\ \hline
\end{tabular}
}
\vspace{-2.0mm}
\label{table:enc}
\end{table}

\section{Conclusion}
\label{sec:conclusion}

We proposed a transformer-based method for multi-view spatio-temporal action recognition~(STAR), called MVAFormer.
To the best of our knowledge, MVAFormer is the first method for the multi-view STAR setting.
In contrast to existing multi-view action recognition methods that recognize only one action for multi-view videos, MVAFormer can recognize spatially and temporally dense actions.
In addition, for effective cooperation among views, MVAFormer utilizes feature maps preserving spatial information and incorporates the novel same view attention and different view attention.
In experiments using the newly collected dataset, MVAFormer outperforms existing single-view and multi-view methods.
We believe that MVAFormer is a significant first step in the multi-view STAR setting.
In future work, we plan to collect a larger dataset with different numbers of views and more scenes to verify whether MVAFormer can achieve a good performance in various environments.



\bibliographystyle{IEEEbib}
\bibliography{strings,refs}

\end{document}